\documentclass[pdflatex,sn-nature]{sn-jnl}

\usepackage{natbib}
\usepackage{etoolbox}

\usepackage{graphicx}%
\usepackage{multirow}%
\usepackage{amsmath,amssymb,amsfonts,mathtools}%
\usepackage{amsthm}%
\usepackage{bm}%

\usepackage[title]{appendix}%
\usepackage{xcolor}%
\usepackage{textcomp}%
\usepackage{manyfoot}%
\usepackage{booktabs}%
\usepackage{listings}%
\usepackage{algorithm}
\usepackage{algpseudocode}
\usepackage{makecell}
\usepackage{lineno}

\theoremstyle{thmstyleone}%
\newtheorem{theorem}{Theorem}
\theoremstyle{thmstyletwo}%

\theoremstyle{thmstylethree}%

\begin{document}
\title[CatBOX: A Categorical--Continuous Bayesian Optimization with Spectral Mixture Kernels for Expedited Catalysis Discovery]
{CatBOX: A Categorical-Continuous Bayesian Optimization with Spectral Mixture Kernels for Accelerated Catalysis Experiments}



\author[1]{\fnm{Changquan} \sur{Zhao}}\email{chomaster@sjtu.edu.cn}

\author[2]{\fnm{Yi} \sur{Zhang}}\email{yz5195@columbia.edu}

\author[1,3]{\fnm{Zhuo} \sur{Li}}\email{li\_zhuo@sjtu.edu.cn}

\author[1]{\fnm{Li} \sur{Jin}}\email{li.jin@sjtu.edu.cn}

\author*[3]{\fnm{Cheng} \sur{Hua}}\email{cheng.hua@sjtu.edu.cn}

\author*[1,4]{\fnm{Yulian} \sur{He}}\email{yulian.he@sjtu.edu.cn}

\affil[1]{\orgdiv{Global College}, \orgname{Shanghai Jiao Tong University}, \orgaddress{\street{800 Dongchuan Rd}, \city{Shanghai}, \postcode{200240}, \country{China}}}

\affil[2]{\orgdiv{Department of Industrial Engineering and Operations Research}, \orgname{Columbia University}, \orgaddress{\street{2960 Broadway}, \city{New York}, \postcode{10027}, \state{NY}, \country{US}}}

\affil[3]{\orgdiv{Antai College of Economics and Management}, \orgname{Shanghai Jiao Tong University}, \orgaddress{\street{1954 Huashan Rd}, \city{Shanghai}, \postcode{200030}, \country{China}}}

\affil[4]{\orgdiv{School of Chemistry and Chemical Engineering}, \orgname{Shanghai Jiao Tong University}, \orgaddress{\street{800 Dongchuan Rd}, \city{Shanghai}, \postcode{200240}, \country{China}}}
\abstract{
Identifying optimal catalyst compositions and reaction conditions is central in catalysis research, yet remains challenging due to the vast multidimensional design spaces encompassing both continuous and categorical parameters. In this work, we present CatBOX, a Bayesian Optimization method for accelerated catalytic experimental design that jointly optimizes categorical and continuous experimental parameters. Our approach introduces a novel spectral mixture kernel that combines the inverse Fourier transform of Gaussian and Cauchy mixtures to provide a flexible representation of the continuous parameter space, capturing both smooth and non-smooth variations. Categorical choices, such as catalyst compositions and support types, are navigated via trust regions based on Hamming distance. For performance evaluation, CatBOX was theoretically verified based on information theory and benchmarked on a series of synthetic functions, achieving more than a 3-fold improvement relative to the best-performing baseline and a 19-fold improvement relative to random search on average across tasks. Additionally, three real-life catalytic experiments, including oxidative coupling of methane, urea-selective catalytic reduction, and direct arylation of imidazoles, were further used for in silico benchmarking, where CatBOX reliably identified top catalyst recipes and reaction conditions with the highest efficiencies in the absence of any a priori knowledge. Finally, we develop an open-source, code-free online platform to facilitate trial deployment in real experimental settings, particularly for self-driving laboratory environments. 
}




\maketitle
\section{Introduction}\label{sec1}

Catalyst design and optimization are central challenges in the development of complex catalytic processes. Whether in catalyst compositions, synthesis parameters, or reaction condition optimization, researchers often face a vast design space comprising thousands to millions of possible candidates~\cite{reymond2010chemical,osolodkinProgressVisualRepresentations2015,medina-francoProgressOpenChemoinformatic2022,opreaChemographyArtNavigating2001}. An exhaustive search in such a high-dimensional space via either the typical intuition-based or model-based practices, or even the advanced high-throughput experimentation~\cite{cai2025synergistic,chang2024high,ozonder2025RapidDiscoveryGraphene} is essentially impractical and resource-demanding. In reality, where experimental budgets are limited, chemists often seek to sample the vast space using factorial design methods such as orthogonal experiments, such that the dimension can be simplified by systematically deconvoluting factorial effects and interactions. While these methods typically demand predefined experimental matrices derived from literature precedents and chemical intuition, they become inefficient and inflexible in high-dimensional or nonlinear design systems, especially when categorical variables are involved, as is often the case for a typical catalyst optimization task, where co-optimization of categorical and continuous parameters is required. 

Alternatively, data-driven optimization approaches have emerged as potential solutions for accelerating catalysis discovery in complex design spaces. Rather than deriving an explicit mechanistic model, data-driven methods excel in inferring structure and correlation based on existing data in an efficient manner. Specifically in experimental design, Bayesian Optimization (BO) stands out as an efficient probabilistic global optimization method especially when only small sample sizes are available~\cite{agarwalDiscoveryEnergyStorage2021,herbolCosteffectiveMaterialsDiscovery2020,ramirez2024accelerated,wahabMachinelearningassistedFabricationBayesian2020,wangBayesianOptimizationChemical2022,xieAccelerateSynthesisMetal2021,zhangBayesianOptimizationMaterials2020}. In brief, BO adopts an adaptive framework by iteratively building a probabilistic surrogate model of the objectives, usually a Gaussian process (GP), and suggesting the most informative experiment to perform next. By balancing exploration of new regions against exploitation of known promising regions, BO can often find high-performing solutions orders of magnitude faster than unguided searches. This approach has been successfully applied to various chemical and materials optimization tasks, particularly suitable for decision-making in autonomous experimentation ~\cite{agarwalDiscoveryEnergyStorage2021,ramirez2024accelerated}. 

Despite its success, applying BO to mixed-parameter optimization problems remains a formidable challenge. Classical BO methods usually assume a continuous, smooth search space; the presence of categorical choices breaks this smoothness and renders gradient-based search for optima difficult~\cite{cuesta2022comparison}. In essence, when categorical parameters are involved, the input space becomes non-differentiable and more complex to model due to the lack of natural ordering of each categorical parameter. To date, a variety of research directions have been explored for mixed-parameter optimization. A common approach is to apply one-hot encoding (OHE) to categorical parameters and treat them as continuous variables. However, the resulting function is only defined on valid one-hot points, violating the GP smoothness assumption. In addition, OHE greatly increases dimensionality, enlarging the search space and making GP modeling and acquisition optimization more difficult. Other common approaches can be broadly categorized into two lines of research. One line focuses on modifying the surrogate model to better accommodate mixed-parameter structures~\cite{bergstra2011algorithms,hase2021gryffin,xian2025unlocking,bliek2021black,oh2019combinatorial}, such as tree-structured models (TPE)~\cite{bergstra2011algorithms}, Bayesian neural network-based surrogates (Gryffin)~\cite{hase2021gryffin}, and alternative functional representations (MVRSM)~\cite{bliek2021black}. A complementary line of work instead concentrates on adapting the acquisition function or the optimization strategy~\cite{ru2020bayesian,wan2021think,hennig2012entropy,hernandez2014predictive}, such as CoCaBO~\cite{ru2020bayesian}, which modified the acquisition process through multi-armed bandit techniques. 

Nevertheless, existing approaches for mixed categorical-continuous BO face intrinsic limitations in surrogate modeling. Tree-based methods and related heuristics provide robust partitioning of the search space but lack a coherent notion of smoothness and uncertainty across continuous dimensions, which often leads to degraded performance in later optimization stages~\cite{hutter2011sequential}. Neural-network-based surrogates, while expressive, do not inherently provide calibrated posterior uncertainty and typically require additional approximation schemes that are unreliable in the small-data, distribution-shifting regimes~\cite{lakshminarayanan2017simple}. In this work, we address these challenges by adopting a GP surrogate equipped with a spectral-density-based kernel. This design enables the model to infer multi-scale and structured patterns in the objective function while retaining principled uncertainty estimates, thereby supporting more effective exploration-exploitation trade-offs in mixed-parameter optimization problems.

We explicitly revisit the design of the GP kernel to enhance the expressive capacity of the surrogate model and its ability to capture correlations across the design space. This is particularly important in mixed categorical-continuous settings, where response surfaces often exhibit complex structures and strong cross-parameter interactions. Motivated by this observation, we propose Categorical-continuous Bayesian Optimization for eXperiments (CatBOX) by modeling mixed parameters within a unified GP surrogate. The key innovation of CatBOX is a composite spectral mixture kernel (SMK) inspired by Bochner’s theorem~\cite{stein1999interpolation} that combines Gaussian and Cauchy spectral components. In this formulation, the Gaussian component captures smooth variations in the response surface, while the heavy-tailed Cauchy component supports the modeling of non-smooth behavior and long-range correlations, allowing the surrogate to represent both gradual trends and sharp transitions in the experimental landscape. This added flexibility in continuous modeling alleviates the burden imposed by categorical parameters: most smooth variability is captured by the continuous kernel, allowing the categorical component to concentrate on the intrinsically combinatorial structure. Such a spectral mixture design leads to improved kernel approximation and is empirically observed to achieve higher marginal log likelihood, faster convergence, and a smaller
optimality gap.

Categorical parameters are modeled using a modified exponentiated Hamming kernel, which is combined with the continuous kernel within a unified Gaussian Process framework. The acquisition function is then optimized through a trust-region-based alternating scheme to enable stable optimization in mixed-parameter spaces. On average across tasks, CatBOX achieves a threefold improvement over the best-performing method and a 19-fold improvement relative to random search. In addition, three in silico catalytic experiments emulated from real-life catalysis research were further implemented for algorithm benchmarking in the more representative experimental spaces, where complexity abruptly arises with intricate physicochemical interactions, experimental noises, and domain-specific constraints. In these cases, CatBOX still reliably identified top catalyst recipes and reaction conditions with the highest efficiencies in the absence of any a priori knowledge. CatBOX is believed to be well-suited for a broad range of complex optimization problems involving mixed categorical and continuous parameters beyond catalysis research, with the benefits becoming more pronounced as problem dimensionality increases. Finally, a code-free online platform is developed to support trial implementations of CatBOX in real research laboratories, particularly those equipped with automated facilities. 


\section{Results}\label{sec2}
\subsection{Overview of CatBOX}

\begin{figure}[!htbp]
\centering
\includegraphics[width=\textwidth]{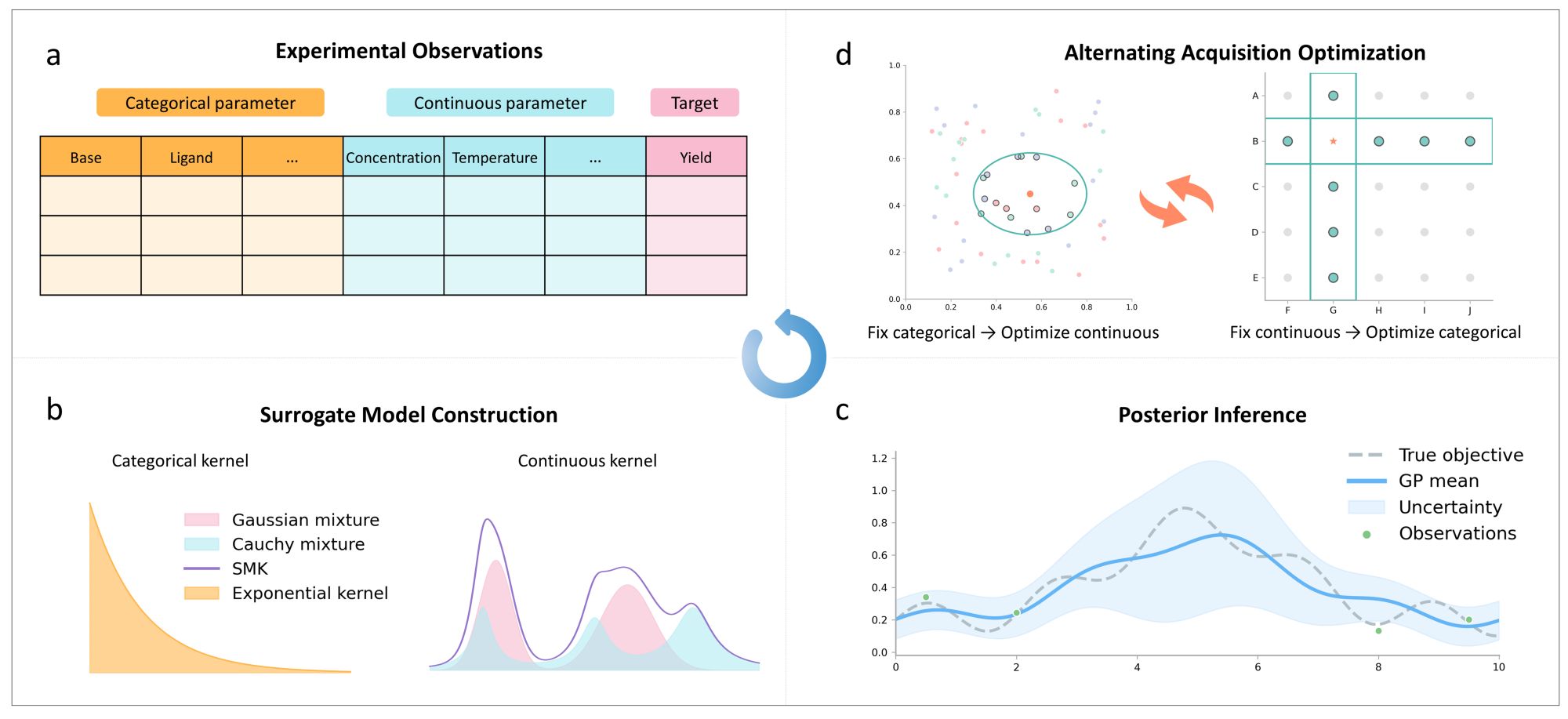}
\caption{| \textbf{CatBOX optimization framework.} \textbf{a,} Existing experimental data are decomposed into categorical and continuous components, together with the target response. \textbf{b,} A surrogate model is constructed using a mixture of Gaussian and Cauchy spectral components for continuous variables and a modified exponentiated Hamming kernel for categorical variables; these are combined via a convex mixture of additive and interaction terms to form a unified Gaussian process. \textbf{c,} Posterior inference of the GP provides the predictive mean and uncertainty, which guide subsequent decisions. \textbf{d,} Optimization alternates between continuous and categorical spaces, with the trust-region radius adaptively adjusted to balance exploration and exploitation.}\label{algorithm}
\end{figure}

The overall workflow of CatBOX is schematically illustrated in Fig. \ref{algorithm}. We want to choose the next experiment when the design space mixes discrete choices (e.g., precursor/support) with continuous knobs (e.g., temperature/concentration). CatBOX fits a Gaussian process surrogate (Supplementary Note 1.1) that predicts both expected performance and uncertainty. The main novelty is the kernel: instead of assuming the response surface has one smoothness level, we model it as a blend of smooth trends and sharp transitions by combining Gaussian- and Cauchy-based spectral components. For discrete choices, we measure similarity by how many categorical fields match, and we learn which categorical fields matter more from data. We then pick the next experiment by maximizing an acquisition score, but we do it within an adaptive neighborhood (trust region), so the search is stable in mixed discrete and continuous spaces. 

Specifically, CatBOX employs a special kernel design to jointly represent the mixed-parameter space. For continuous parameters, CatBOX uses an SMK (Eq.~\eqref{eq:cont-kernel}) composed of Gaussian Spectral Mixtures (GSM) and Cauchy Spectral Mixtures (CSM). Intuitively, the Gaussian component is well-suited for capturing smooth and local trends, such as the gradual performance variations induced by small changes in experimental conditions. In contrast, the Cauchy component, which has heavier tails, allows the model to capture more abrupt, non-smooth behavior and long-range correlations, such as sudden performance shifts caused by subtle experimental parameter changes, as is the case for the observation of a light-off temperature point in catalytic combustion reactions~\cite{yan2025optimizing}, or the high sensitivity of chain propagation on flow conditions in Fischer-Tropsch syntheses~\cite{jiang2025ambient}. The weights of each mixture component are adaptively learned and updated from the observed data by maximizing the marginal log likelihood (MLL). By combining these two components, the surrogate model can flexibly represent both gradual variations and sharp transitions that commonly arise in complex experimental landscapes, while conventional kernels, such as the Radial Basis Function or Matérn kernels (Supplementary Note 1.2), assume a fixed structure that struggles to fully capture the complexities of high-dimensional objective functions. The mixed categorical and continuous variables create additional challenges for these kernels.  A more detailed technical discussion on the comparison of different kernel design strategies can be found in Supplementary Note 2. The design of the proposed kernel stems from Bochner’s Theorem (Supplementary Note 3.1), which provides a foundational result in characterizing stationary positive definite kernels in terms of their spectral representations. We further provide rigorous theoretical guarantees, deriving information gain and regret bounds via information-theoretic analysis (Supplementary Notes 3.2–3.3). 

Building on this analysis, we examine the ability of different kernels to approximate standard kernels and their combinations, and find that the proposed method provides a more effective approximation than either CSM or GSM alone, as well as other baseline kernels (Supplementary Note 3.4). Empirical results further show that it achieves improved relative performance and reduced optimality gaps compared with conventional kernels across a range of optimization tasks with various mixing ratios (Supplementary Note 3.5-3.6), with the advantages becoming more pronounced as problem complexity increases, reflecting enhanced approximation capacity and adaptability. 

Furthermore, categorical parameters are modeled using a weighted exponential Hamming kernel (Eq.~\eqref{eq:cate-kernel}). Unlike the standard Hamming kernel, which assigns equal importance to all categorical features, the kernel used in CatBOX allows different categorical features to contribute unequally to the overall similarity. The weight associated with each categorical feature is estimated from the observed data, reflecting its relative influence on the objective function. This data-driven weighting scheme enables the surrogate model to capture the differing importance of categorical factors, rather than treating all categorical choices as equally informative, as in OHE-based methods. 

The kernels for continuous and categorical parameters are combined through a convex mixture of interaction and additive components in Eq.~\eqref{eq:comp-kernel} to form a unified covariance structure. This unified formulation enables CatBOX to simultaneously learn the effects of continuous and categorical parameters, as well as their interactions. 

After updating the posterior distribution of GP, CatBOX uses an acquisition function to balance exploration and exploitation. In the following discussion, we adopt the Expected Improvement (EI) acquisition function, which measures the expected improvement over the current best observation under the GP posterior and naturally balances exploration and exploitation by jointly accounting for the predicted mean and uncertainty. Other acquisition functions, including Upper Confidence Bound (UCB) and Probability of Improvement (PI), are also evaluated and exhibit consistently strong performance across a range of optimization tasks (Supplementary Note 3.7). 

Finally, the optimization is carried out using a trust-region-based alternating optimization scheme, in which continuous parameters are optimized within a local trust region, while categorical parameters remain unchanged and vice versa (Supplementary Note 4). This strategy enables stable and efficient exploration of mixed categorical-continuous spaces while reducing the risk of premature convergence to local optima. The selected candidate is subsequently evaluated using the true objective function, and the resulting observation is incorporated into the dataset to update the surrogate model. This process is repeated until the evaluation budget is exhausted or convergence criteria are met, yielding a sequence of increasingly improved solutions. The pseudocode of CatBOX is provided in Supplementary Note 5.

\subsection{Synthetic Function Benchmarks}

We first employ a suite of synthetic functions as a controlled and flexible testbed for benchmarking CatBOX against state-of-the-art optimization algorithms. Synthetic functions allow the dimensionality of both categorical and continuous parameters to be adjusted arbitrarily, enabling systematic evaluation of scalability, robustness, and convergence behavior across a wide range of problem settings. This flexibility makes them particularly valuable for isolating algorithmic properties that are difficult to assess in realistic experimental scenarios. 

For clarity and illustrative purposes, the Ackley function (Eq. \eqref{eq: ackley}) is used as a representative example to demonstrate the optimization workflow here. More results of synthetic benchmark functions, including Griewank, Rosenbrock, and Schwefel, are shown in Fig. S8-S10 in Supplementary Note 6. 
\begin{equation}
f_{Ackley}(\bm{x}) = -a \exp\!\left(-b \sqrt{\frac{1}{d}\sum_{i=1}^{d} x_i^2}\right)
               - \exp\!\left(\frac{1}{d}\sum_{i=1}^{d} \cos(c x_i)\right)
               + a + e, 
\label{eq: ackley}
\end{equation}
where $a$, $b$, and $c$ are parameters of the Ackley function and are set as $a=20$, $b=0.2$, and $c=2\pi$ in this work. The parameters $a$, $b$, and $c$ control the depth of the global basin, the steepness of the landscape, and the frequency of local oscillations in the Ackley function, respectively. The parameter $d$ denotes the dimensionality of the input parameters that can be tuned for dimensional analysis, and $e$ is the base of the natural logarithm.

\begin{figure}[!htbp]
\centering
\includegraphics[width=\textwidth]{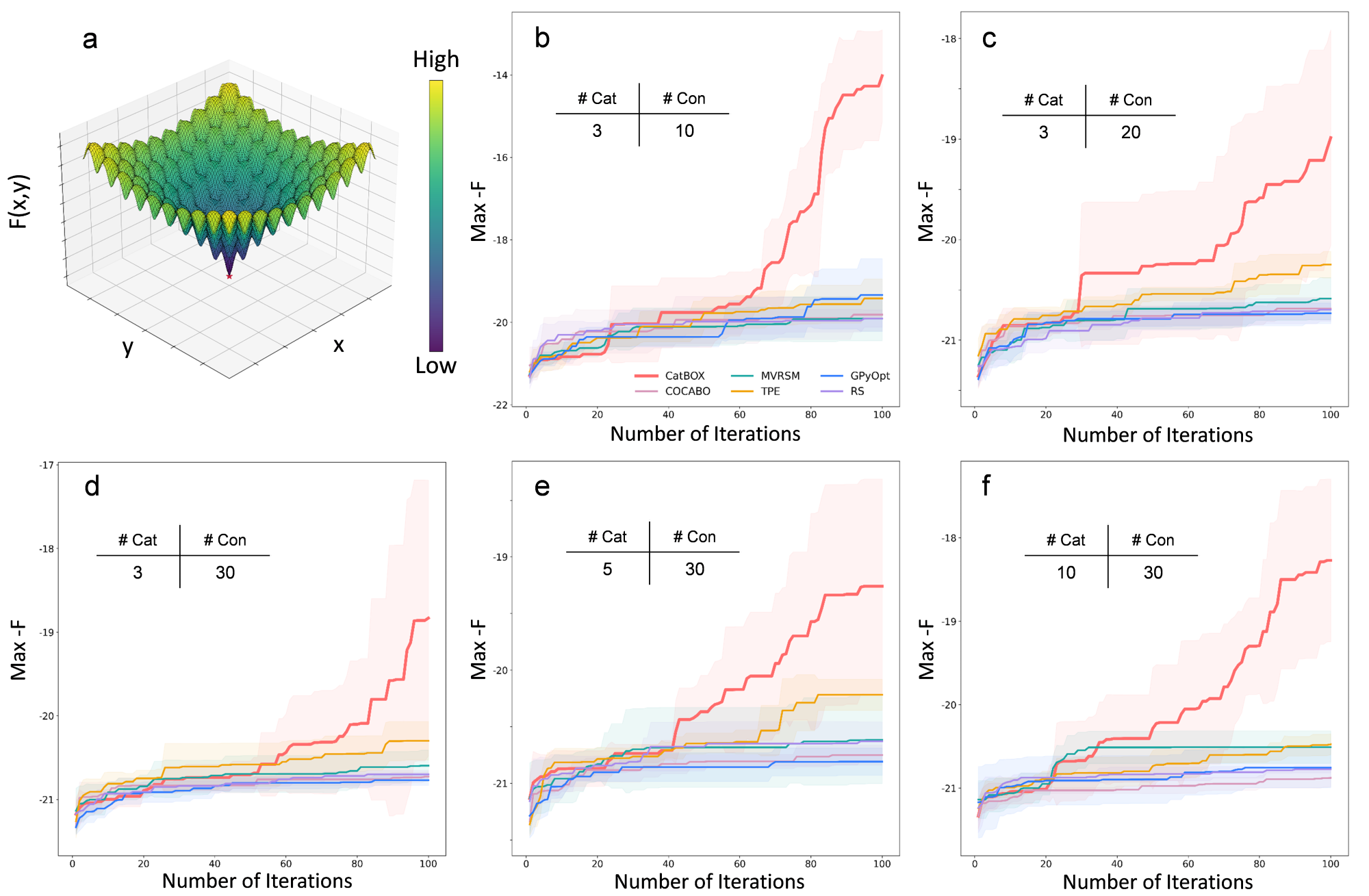}
\caption{| \textbf{Mixed-variable optimization on the Ackley function.} \textbf{a,} Illustration of the Ackley function. 
\textbf{b-f,} Optimization results of different algorithms on mixed categorical-continuous Ackley problems with varying dimensional configurations, where the number of categorical (\# Cat) and continuous (\# Con) parameters is indicated in each panel. Solid lines denote the mean incumbent objective value over 5 repeated runs, while shaded regions indicate the corresponding standard deviation.}
\label{Ackley_fig}
\end{figure}

The Ackley function is a highly non-convex and multi-modal benchmark (Fig. \ref{Ackley_fig}a) with an almost flat outer region and a very narrow global optimum basin. It is commonly used to assess the global exploration capability of optimization algorithms in the presence of numerous local minima traps. As the dimensionality of both continuous and categorical parameters increases (Fig. \ref{Ackley_fig}b-f), CatBOX consistently demonstrates outstanding performance and shows a reduced tendency toward premature convergence compared to baseline methods. This tendency was also observed for other test functions, as detailed in Supplementary Note 6.

\subsection{In silico Catalytic Experiments}


While synthetic functions are useful for benchmarking, they do not capture the full complexity of realistic chemical reactions, which involve noise, physicochemical interactions, and domain-specific constraints. To address this, we adopt in silico experimentation, where virtual reaction spaces are constructed from real chemical reaction data using an AutoML framework (Supplementary Note 7). This approach enables efficient evaluation of CatBOX against state-of-the-art experimental planning strategies~\cite{hase2021olympus,burger2020mobile,li2024sequential}. Three in silico catalytic reactions with complex mixed categorical and continuous parameter spaces are considered below.

\subsubsection{Oxidative Coupling of Methane (OCM)}
\begin{figure}[htbp]
\centering
\includegraphics[width=\textwidth]{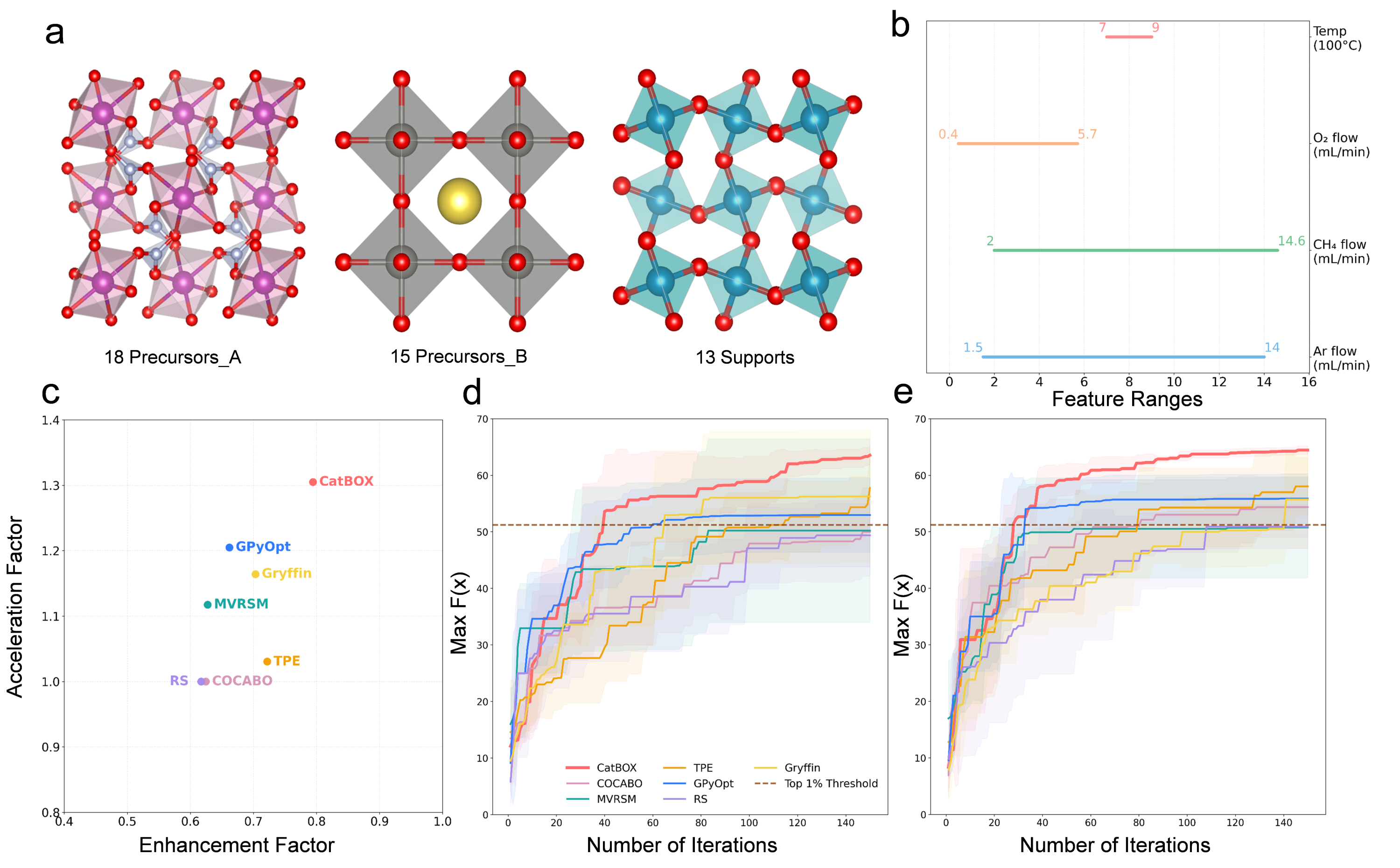}
\caption{| \textbf{Comparative study for OCM reaction system.} \textbf{a,} Categorical catalyst components. Color scheme: Mn (purple), N (silver), O (red), Na (yellow), W (gray), and Si (blue). \textbf{b,} Continuous parameters and their ranges. \textbf{c,} Acceleration Factor and Enhancement Factor of different optimization methods. \textbf{d,} Comparison of optimization performance across different algorithms. Solid lines indicate mean values over 5 repeated runs, and shaded regions represent $\pm 1$ standard deviation. \textbf{e,} Comparison of optimization performance across different algorithms in noisy conditions.}\label{fig: OCM}
\end{figure}

Oxidative coupling of methane (OCM) is an important class of catalytic reactions where methane can be upgraded to value-added coupled products such as ethane (C\textsubscript{2}H\textsubscript{6}) and ethylene (C\textsubscript{2}H\textsubscript{4}). Albeit economically appealing, the commercial deployment of this reaction has been largely limited by its conversion-selectivity trade-off that arises from both the high chemical inertness of methane and its oriented transformation to desired products. Catalyst development of multi-functionalities is central to breaking the trade-off, while conventional experimental planning practices are inefficient since a multidimensional and mixed-parameter space spanning catalyst features and reaction conditions will be involved. For instance, an optimization task of the OCM process will typically involve the selection of different metals (M1, M2, M3) and support categories (S) in the form of M1-M2\textsubscript{1-2}M3O\textsubscript{4}/S, as well as temperature, pressure, CH\textsubscript{4}:O\textsubscript{2} feed-in ratio, etc. In this sense, an in silico virtual experimentation is conducted to compare CatBOX to other experimental planning strategies for the optimization of complex OCM catalyst recipes and process conditions~\cite{feng2023bayesian,burger2020mobile}. 

The virtual OCM experimental space was first constructed by fitting a realistic experimental dataset from high-throughput experiments (HTE). This dataset includes 12,708 realistic experimental performance data points on 59 different catalysts with varying metal/support categories and 216 different reaction conditions~\cite{nguyen2019high}. The optimization target of this reaction is to maximize the yields (\%) for C2 products, while with higher rewards for ethylene (C\textsubscript{2}H\textsubscript{4}) over ethane (C\textsubscript{2}H\textsubscript{6}) and higher penalties for side-product carbon dioxide (CO\textsubscript{2}) that can be formed from either methane complete combustion or coke oxidation over carbon monoxide (CO). In this sense, we have designed the objective function as follows:  
\begin{equation}
    \label{target}
  Target (\%) = Conv_{CH_4}\frac{2S_{C_2H_4} + S_{C_2H_6}}{2S_{CO_2} + S_{CO}},
\end{equation}
where $Conv_{CH_4}$ represents the conversion of $CH_4$ (\%), $S$ represents selectivity (\%). The coefficient is designed to reflect the differently weighted rewards/penalties towards each product.

All fitting procedures in this work were carried out using AutoGluon-Tabular 1.2~\cite{erickson2020autogluon}, which automates the selection of the most appropriate machine learning models and (hyper)parameter settings from numerous models available without human intervention (Supplementary Note 7). The best model to implicitly describe the OCM experimental space was found to be the WeightedEnsemble model with the lowest mean absolute error (MAE) of 2.15\% on the test set.


A total of four categorical parameters (M1, M2, M3, S) and 4 continuous parameters are involved for the OCM optimization task (Fig. \ref{fig: OCM}a,b). To ensure data reliability, we consolidated the precursor formulation into a single categorical variable (e.g., Mn\textsubscript{0.371}|Na\textsubscript{0.370}|W\textsubscript{0.185}|SiC) to avoid wild extrapolation. In this implicit virtual chemical space, we conducted five independent optimization runs for all state-of-the-art algorithms, including Gryffin~\cite{hase2021gryffin}, which was specifically designed for tackling mixed-parameter optimization in chemical reactions, MVRSM~\cite{bliek2021black}, TPE~\cite{bergstra2011algorithms}, CoCaBO~\cite{ru2020bayesian}, and a standard software package called GPyOpt, each with 150 iterations. For each algorithm, 20 randomly-sampled data points were used as the initial inputs, analogous to the worst-case scenario where an experimentalist has no a priori knowledge of a specific chemical process. The results show that CatBOX achieves a higher Enhancement Factor (EF) and Acceleration Factor (AF, Supplementary Note 8) than all other algorithms compared (Fig. \ref{fig: OCM}c). During the early exploration phase (iterations 0-30), all algorithms have induced a performance increment while at different paces, with CatBOX, GPyOpt, and MVRSM outperforming the rest, reaching the top 3.4\% performance targets in the realistic HTE distribution (Fig. \ref{fig: OCM}d). Remarkably, further improving the target performance turned out to be challenging for all the algorithms except for CatBOX, which evolved continuously not only at a faster acceleration rate but also converged to the top 0.58\% with a greater EF within a total of 50 iterations. Such high-performance data points correspond to an extremely low occurrence probability in the realistic HTE dataset. 

Accounting for the errors in realistic experiments, we applied Gaussian noise with a mean of 0 and standard deviations of 5\% of the global optimum target (69.9) to all data points targeted at each optimization iteration (Fig. \ref{fig: OCM}e). The same noise configuration was consistently maintained throughout all experiments. Under the noisy conditions, all methods exhibit increased uncertainty and slower early-stage convergence. Nevertheless, CatBOX consistently maintains the fastest improvement rate and the highest incumbent values throughout the optimization, indicating that its performance advantage is robust to a moderate level of noise that is commonly found in realistic chemical experiments.

\subsubsection{Selective Catalytic Reduction (SCR)}

\begin{figure}[htbp]
\centering
\includegraphics[width=\textwidth]{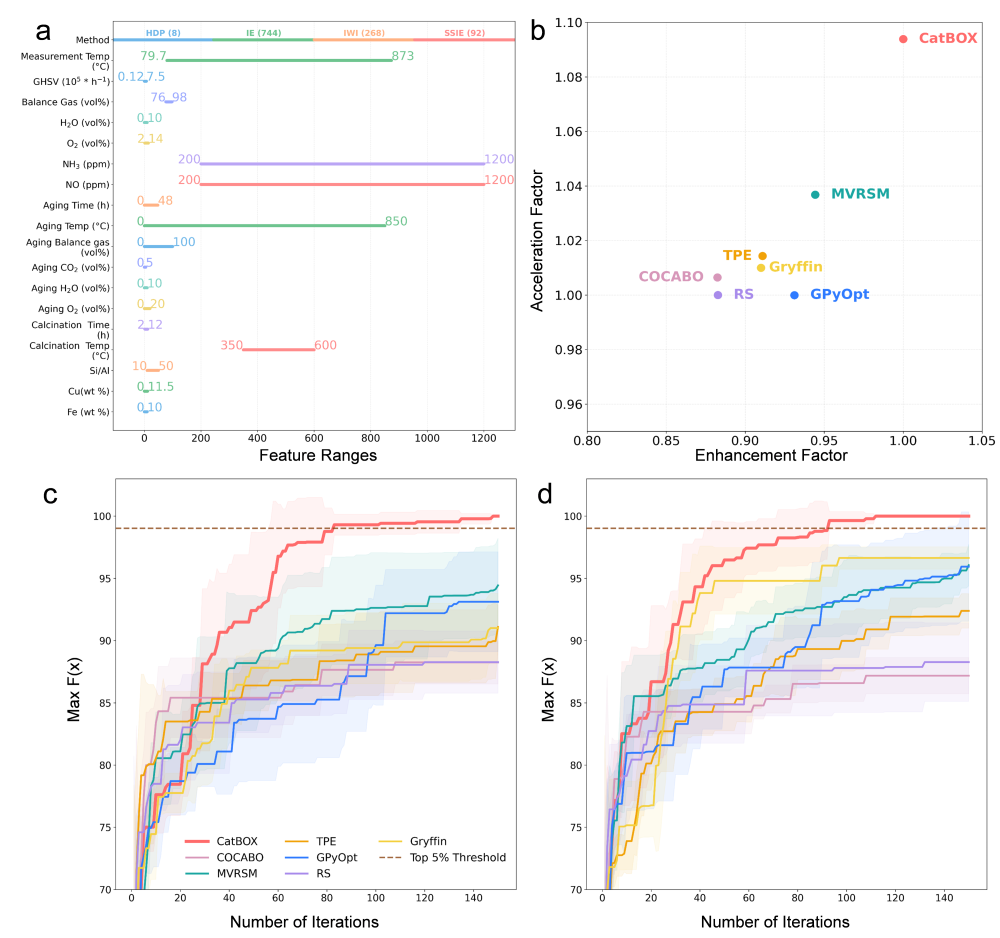}
\caption{| \textbf{Comparative study for SCR reaction system.} \textbf{a,} Parameter ranges defining the optimization search space. Abbreviations: HDP, homogeneous deposition precipitation; IE, ion exchange;
IWI, incipient wetness impregnation; SSIE, solid-state ion exchange. Numbers in parentheses denote the number of data points for each method. \textbf{b,} EF and AF analysis for different optimization methods in the noise-free setting.
\textbf{c,} Optimization results under noise-free conditions, showing the evolution of the incumbent solution over iterations.
\textbf{d,} Optimization results with observational noise added to the objective evaluations.}\label{fig: SCR}
\end{figure}

$NO_x$ is a class of pollutant gases typically emitted by diesel vehicles. In exhaust gas treatment, most approaches introduce reducing agents like ammonia or urea to selectively convert $NO_x$ into harmless nitrogen gas under the aid of catalysts, in this case zeolites, of which the catalytic performance is governed by multiple structural factors that are highly sensitive to a plethora of synthesis parameters. This dataset was collected from literature reports~\cite{bae2022data}, including 1,112 data points over three types of zeolite-based catalysts ($\beta$, ZSM-5, and SSZ-13) with the optimization target being maximizing the conversion (\%) over numerous synthesis and reaction variable combinations. Following the same procedures,  we first emulated the virtual reaction space using AutoML, the resulting WeightedEnsemble model implicitly learned a chemical space described by 19 experimental parameters, including one categorical variable that differentiates the catalyst synthesis methods (Fig. \ref{fig: SCR}a). The final model achieved a mean absolute error (MAE) of 4.59\%.

In the noise-free setting (Fig. \ref{fig: SCR}b, c), CatBOX shows a markedly faster improvement and converges to near-optimal performance with significantly higher EF and AF values than the competing methods. The accelerated convergence indicates an efficient balance between exploration and exploitation, enabling CatBOX to be the only algorithm to identify the top 5\% high-performance regions in the chemical space. In contrast, other algorithms exhibit more gradual performance gains and larger variability across repeated runs, suggesting a stronger dependence on initialization and less effective space contraction.

When observational noise is introduced (Fig. \ref{fig: SCR}d), the overall performance gap between CatBOX and other optimization methods becomes less pronounced. This behavior arises because noise can positively bias the measured performance of individual samples, and the incumbent solution directly retains such occasional overestimations. As a result, the apparent upper bounds of different methods are numerically elevated, causing their incumbent trajectories to converge and reducing the discriminative power of this metric. Despite this “compression” effect, CatBOX consistently maintains a favorable position throughout the optimization process, exhibiting stable behaviors in both convergence and final performance. 

\subsubsection{Direct Arylation of Imidazoles (DAr)}

Apart from heterogeneous catalytic processes, we also considered an important homogeneous catalytic reaction, direct arylation of imidazoles (DAr), a representative palladium-catalyzed C-H functionalization reaction that enables the direct formation of C-C bonds on heteroaromatic imidazole scaffolds without prior substrate prefunctionalization. Owing to its high atom economy and relevance to pharmaceutical synthesis, DAr has attracted sustained interest as both a synthetic method and a model system for studying reaction optimization. However, the reaction outcome is highly sensitive to a multidimensional combination of categorical and continuous parameters, including ligand identity, base, solvent, temperature, and concentration, making systematic optimization particularly challenging.

\begin{figure}[!htbp]
\centering
\includegraphics[width=\textwidth]{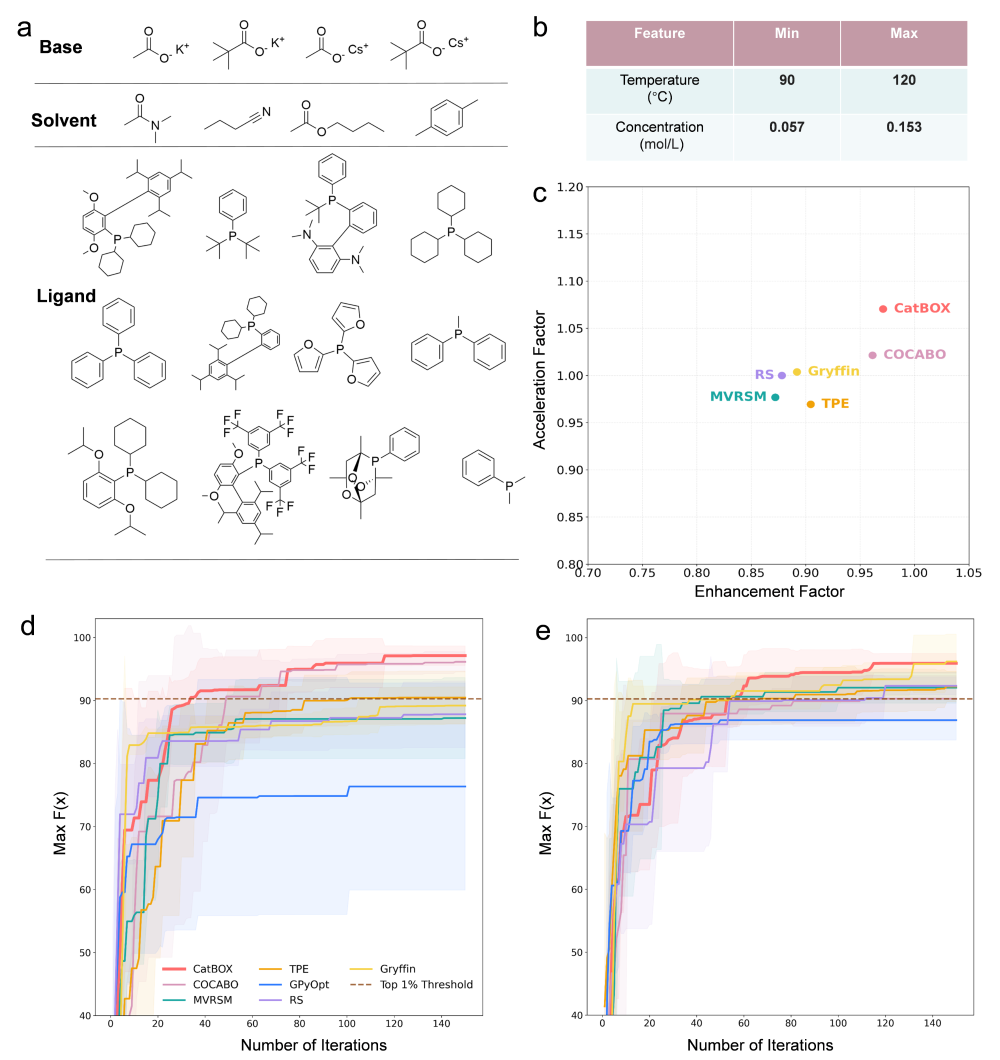}
\caption{| \textbf{Comparative study for DAr reaction system.} 
\textbf{a,} Categorical reaction components considered in the DAr optimization space, including bases, solvents, and ligands.
\textbf{b,} Continuous reaction parameters (temperature and concentration of substrate~\cite{shields2021bayesian}) and their corresponding ranges
\textbf{c,} EF and AF analysis for different optimization methods in the noise-free setting.
\textbf{d,} Optimization results under noise-free conditions, showing the evolution of the incumbent solution over iterations.
\textbf{e,} Optimization results with observational noise added to the objective evaluations.}\label{fig: DAr}
\end{figure}

For establishing the in silico reaction space, we adopt the DAr dataset reported by Shields et al.~\cite{shields2021bayesian} via systematic HTE. The dataset exhaustively covers 1,728 distinct reaction conditions, defined by combinations of 12 ligands, 4 bases, 4 solvents, and 3 levels of temperatures and concentrations (Fig. \ref{fig: DAr}a, b), with the maximization of reaction yield (\%) serving as the optimization objective. In the entire dataset, approximately one-third of the data points have a conversion rate of zero. This is because the DAr reaction space exhibits strong clustering among high-performing recipes. Once one or two categories strongly correlated with performance are identified, recipes with similar categorical configurations tend to yield comparable performance. Using the same experimental setup and model configuration as in the preceding analysis, the AutoML regression model yielded an MAE of 3.9\%.

In the DAr reaction system, CatBOX exhibits stable and efficient optimization behavior under both noise-free (Fig. \ref{fig: DAr}c, d) and noisy settings (Fig. \ref{fig: DAr}e). Under noise-free conditions, CatBOX is able to rapidly identify the top 1\% recipes within just 40 iterations and consistently exploit high-yield regions in later stages, demonstrating superior convergence speed and trajectory stability. Upon introducing observational noise, the performance metric is numerically influenced by noise-induced overestimation, and the performance differences among methods become compressed, while CatBOX consistently remains on the upper envelope of performance and retains clear advantages in both optimization efficiency and stability. It is worth noting that random search (RS) serves as a relatively strong baseline in this problem, primarily due to the structural characteristics of the DAr reaction space, where high-yield regions are relatively dense and dominated by a small number of critical categorical parameters such as ligand and base identity. As a result, for a highly clustered design space like DAr, the margin for improvement between different optimization algorithms might be inherently limited.

\subsubsection{Optimization Path Analysis in Categorical Parameters}

\begin{figure}[!htbp]
\centering
\includegraphics[width=\textwidth]{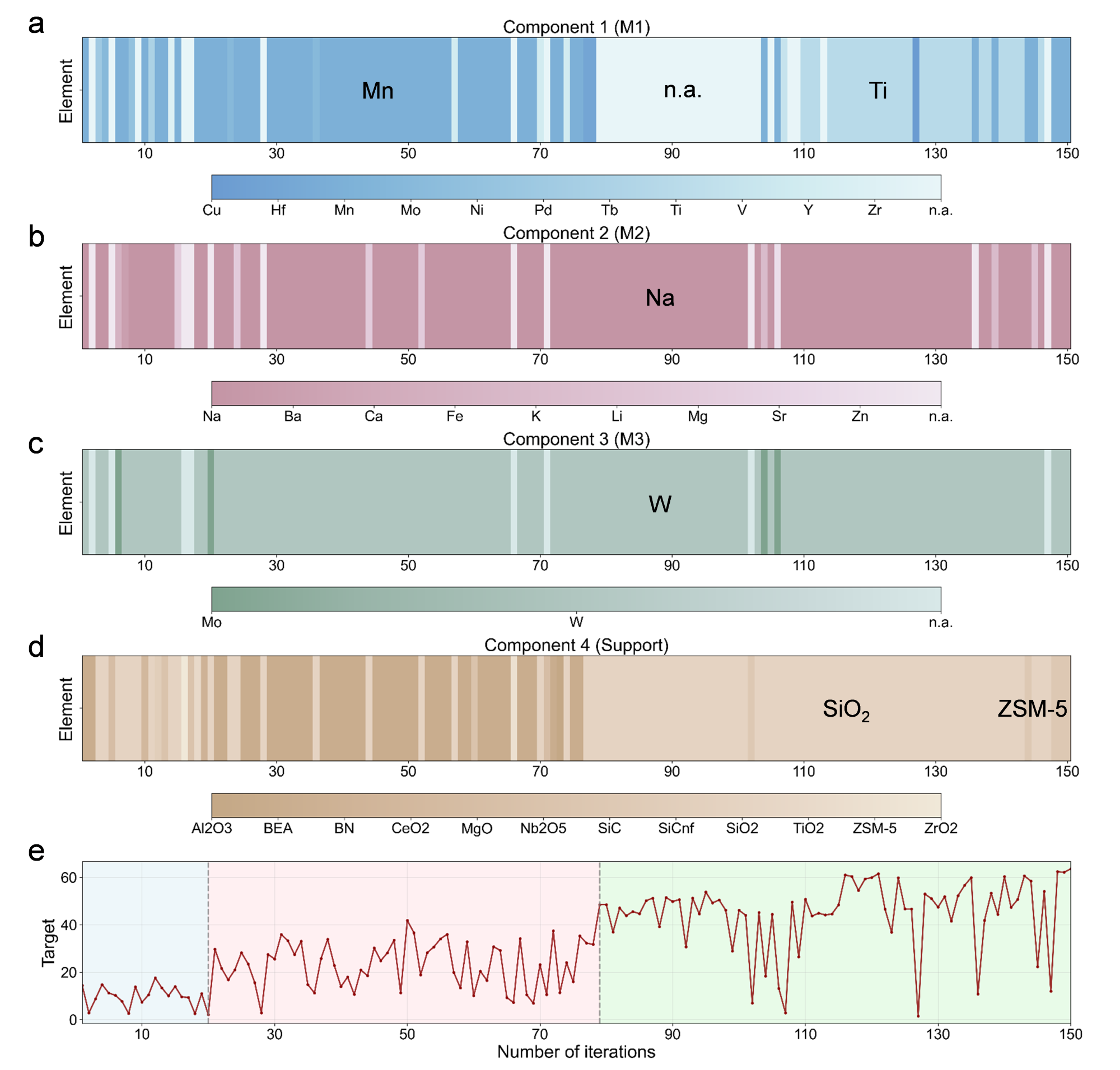}
\caption{| \textbf{Path analysis of categorical parameters.}
\textbf{a--d,} Evolution of categorical selections over iterations for Component 1 (M1), Component 2 (M2), Component 3 (M3), and Component 4 (Support). Each vertical stripe indicates the category selected at a given iteration, with colors denoting different categorical options.
\textbf{e,} Target value as a function of iteration number. The discrete configuration at each iteration corresponds to the selections shown in panels \textbf{a--d}. }\label{categorical_analysis}
\end{figure}

To unveil the decision-making strategies of CatBOX, we further show the evolution of an optimization path in the OCM problem. In the early optimization stage (20-30), CatBOX showed a clear preference in exploration over exploitation in categorical spaces. These early explorations ensure a preliminary understanding of the design space by CatBOX prior to enhancing the exploitation weights. After approximately 30 iterations, the model rapidly identifies Mn as an effective promoter and Na\textsubscript{2}WO\textsubscript{4} as the active catalyst at an early stage. This corresponds to the state-of-the-art OCM catalyst recipes that took decades to formulate since the first discovery of the OCM reaction in 1982~\cite{keller1982synthesis}. While the optimization did not stop but continued to further explore the design space, it subsequently discovered that formulations without a promoter or with Ti as the promoter can also yield favorable performance. This finding is highly consistent with prior experimental studies reporting Ti as an effective promoter capable of enhancing C\textsubscript{2} yield~\cite{nguyen2019high}. While these configurations only correspond to local optima, within a few more iterations, CatBOX is able to escape from these local optima and converges to the Mn promoted Na\textsubscript{2}WO\textsubscript{4} with ZSM-5 or SiO\textsubscript{2} as the optimal support formulation, corresponding to state-of-the-art recipes. The reveal of the decision path during optimization further confirms the strong chemical space approximation ability of CatBOX regardless of its high-dimensional, mixed-variables, and noisy nature. Additional decision path analyses on SCR and DAr can be found in supplementary Note 9.

\section{Discussion}

In this study, we develop a composite kernel design and Bayesian optimization framework for mixed-variable experimental design in catalysis. The proposed approach incorporates a mixed spectral continuous kernel constructed from the inverse Fourier transforms of Gaussian and Cauchy mixtures based on Bochner’s theorem, together with a modified categorical exponential kernel integrated with a trust-region strategy. The mixed spectral kernel combines Gaussian and Cauchy components: the Gaussian component captures smooth variations with exponential decay in the squared distance, while the Cauchy component employs exponential decay in distance to better accommodate non-smooth behavior and outliers. This structure enhances modeling flexibility for complex correlations in high-dimensional mixed-variable spaces and contributes to the strong empirical performance of CatBOX across synthetic benchmarks and realistic catalytic studies. 

CatBOX adapts to the underlying response surface as data are collected by dynamically updating both the weights of the spectral mixture components and the automatic relevance determination parameters associated with each Hamming distance. This data-driven adaptation allows the surrogate model to adjust its inductive bias over time, in contrast to standard Bayesian optimization methods that rely on fixed kernels and therefore require strong prior assumptions about the functional form of the objective. By learning kernel structure from observations, CatBOX remains flexible across both smooth and non-smooth regions of the search space and supports effective exploration even in the absence of a priori knowledge.

The optimization logic of CatBOX closely reflects the decision-making process commonly adopted by experimental scientists. It follows an explore-first, exploit-later strategy, enabling systematic exploration of the design space before focusing on high-potential regions. When convergence criteria are approached, CatBOX further introduces verification and targeted re-exploration steps to assess the reliability of the current solutions, thereby enhancing robustness and reducing the risk of premature convergence to local optima.


\section{Methods}
\subsection{Problem Statement}
Consider the problem of maximizing an unknown objective function $f$, formulated as:
\begin{equation*}
    \bm{x}^* = \arg\max_{\bm{x} \in \mathcal{X}} f(\bm{x}),
\end{equation*}
where the search space $\mathcal{X}=\mathcal{X}_u \times \mathcal{X}_c$ contains both categorical and continuous variables. The categorical space is defined as $\mathcal{X}_u \subset \mathbb{D}_1 \times \cdots \times \mathbb{D}_U$, where each $\mathbb{D}_i$ is a finite set associated with categorical variable $i$. The continuous space is $\mathcal{X}_c \subset \mathbb{R}^d$. Each input is written as $x=(x_u,x_c)$, with $x_u \in \mathcal{X}_u$ and $x_c \in \mathcal{X}_c$. We write each input as $\bm{x} = (\bm{x}_u, \bm{x}_c)$, where $\bm{x}_u \in \mathcal{X}_u$ and $\bm{x}_c \in \mathcal{X}_c$. The solution $\bm{x}^*$ corresponds to a global maximizer of $f$ over the mixed domain. Throughout this paper and the supplementary information, we use 
$d$ to denote the dimensionality of the continuous subspace only. The number of categorical variables is denoted separately by $U$.


\subsection{Continuous Search Space: Spectral Mixture Kernel} 
\label{ssec:spectral_kernel}
A kernel defines how similar two experiments are expected to be and therefore shapes the structure of the Gaussian process surrogate in Bayesian optimization. The choice of kernel plays a central role in BO performance, as it determines how the underlying response surface is modeled. Most existing BO solvers rely on conventional kernels such as the squared exponential, rational quadratic, and Matérn kernels, which typically assume a single smoothness scale. Although simple and computationally convenient, these kernels often struggle to capture the complexity observed in practical applications~\cite{wilson2013gaussian}. In catalysis, for example, responses may vary smoothly in some regions while exhibiting sharp transitions near critical thresholds, violating the single-scale smoothness assumption.

To address this limitation, we propose a spectral mixture kernel motivated by Bochner’s theorem~\cite{stein1999interpolation}, which states that every stationary kernel can be represented as the Fourier transform of a finite positive spectral measure. A spectral mixture kernel models the unknown response as a weighted sum of wave-like components at different characteristic frequencies, allowing multiple smoothness scales to coexist. By learning the mixture weights from data, the model can flexibly adapt its smoothness structure, enabling it to capture both gradual trends and abrupt changes in experimental landscapes.

\begin{theorem}[Bochner's Theorem~\cite{stein1999interpolation}]
\label{thm:BochnerTHM}
A complex-valued function \(k\) on \( \mathbb{R}^d \) is the kernel of a weakly stationary, mean square continuous complex-valued random process on \( \mathbb{R}^d \) if and only if it can be represented as
\begin{equation*}
        k(\bm{\tau}) 
    = \int_{\mathbb{R}^d} 
    e^{2\pi i \bm{s}^{\top}\bm{\tau}}
    \, d\psi(\bm{s}),
\end{equation*}
where \( \psi \) is a positive finite Borel measure on \( \mathbb{R}^d\), $\bm{s} \in \mathbb{R}^d$ is a frequency variable, and $\bm{\tau} = \bm{x}_c - \bm{x}_c'$ with $\bm{x}_c, \bm{x}_c' \in \mathbb{R}^d$.
\end{theorem}
In other words, the Fourier transform of such a measure yields a stationary covariance function~\cite{stein1999interpolation}. The measure \(\psi\) is called the \textit{spectral measure} of \(k\). If \( \psi \) has a density \( S \), then \( S \) is referred to as the \textit{spectral density} or \textit{power spectrum} of \( k \). The covariance function \( k \) and the 
spectral density \( S \) form a Fourier pair:
\begin{equation}
\label{kernelEQ}
k(\bm{\tau}) 
= \int_{\mathbb{R}^d} 
S(\bm{s}) e^{2\pi i \bm{s}^{\top}\bm{\tau}} d\bm{s},
\qquad
S(\bm{s}) 
= \int_{\mathbb{R}^d} 
k(\bm{\tau}) e^{-2\pi i \bm{s}^{\top}\bm{\tau}} d\bm{\tau}.
\end{equation}

\subsubsection{Gaussian Spectral Density}
A natural choice for constructing a space of stationary kernels is to use a mixture of Gaussian distributions~\cite{wilson2013gaussian} to represent the spectral density $S_g(\bm{s})$:
\begin{equation}
\label{eq:dstGaussian}
\phi_{ g}(\bm{s}) = \sum_{q=1}^{Q_g} w_q^g \mathcal{N}(\bm{s}; \bm{\mu}_q, \bm{\Sigma}_q), \qquad S_g(\bm{s}) = \frac{\phi_g(\bm{s}) + \phi_g(-\bm{s})}{2},
\end{equation}
where the construction of the spectral density $S_g$ ensures symmetry, and the weights \( w_q^g \) determine the contribution of each of the $Q_g$ components. The covariance matrix $\bm{\Sigma}_q$ controls the spectral bandwidth and thus determines the smoothness of the function in the input space, whereas the mean vector $\bm{\mu}_q$ specifies the oscillation frequency of the corresponding kernel component. The symmetry ensures the corresponding kernel $k(\bm{\tau})$ to be real-valued. By taking the inverse Fourier transform in Eq.~\eqref{kernelEQ}, the resulting spectral mixture kernel induced by Gaussian distributions is given by:
\begin{align}
\label{eq:gsm}
k_{g}(\bm{\tau})
&= \int_{\mathbb{R}^d} S_g(\bm{s})\, e^{2\pi i \bm{s}^{\top}\bm{\tau}}\, d\bm{s} \nonumber\\
&= \frac{1}{2}\int_{\mathbb{R}^d}\!\big(\phi_g(\bm{s})+\phi_g(-\bm{s})\big)\, e^{2\pi i \bm{s}^{\top}\bm{\tau}}\, d\bm{s} \nonumber\\
&= \frac{1}{2}\sum_{q=1}^{Q_g} w_q^g \int_{\mathbb{R}^d}\!\mathcal{N}(\bm{s};\bm{\mu}_q,\bm{\Sigma}_q)\, e^{2\pi i \bm{s}^{\top}\bm{\tau}}\, d\bm{s}
 +\frac{1}{2}\sum_{q=1}^{Q_g} w_q^g \int_{\mathbb{R}^d}\!\mathcal{N}(\bm{s};-\bm{\mu}_q,\bm{\Sigma}_q)\, e^{2\pi i \bm{s}^{\top}\bm{\tau}}\, d\bm{s} \nonumber\\
&= \sum_{q=1}^{Q_g} w_q^g\,
e^{-2\pi^2 \bm{\tau}^{\top}\bm{\Sigma}_q \bm{\tau}}
\cos\!\left(2\pi \bm{\tau}^{\top}\bm{\mu}_q\right).
\end{align}
Inspecting Eq.~\eqref{eq:gsm}, we observe that the covariance function induced by a Gaussian mixture spectral density is infinitely differentiable. However, this choice may generate overly smooth sample paths~\cite{garnett2023bayesian}.

\subsubsection{Cauchy Spectral Density}
To address the issue of overly smooth sample paths in GSM, we introduce a different family of distributions, the  Cauchy distribution $\mathcal{C}(\bm{s}; \bm{\eta}_q, \bm{\gamma}_q)$
to construct a class of continuous but finitely differentiable covariance functions, where $\bm{\eta}_q$ is central frequency parameter and $\bm{\gamma}_q$ is scale parameter. 

Following a derivation analogous to that of the GSM, we construct the spectral density in a symmetric manner to ensure that the resulting covariance function is real-valued. Specifically, we begin with a mixture of Cauchy spectral components and then enforce symmetry by averaging the density with its reflection. Applying the inverse Fourier transform in Eq.~\eqref{kernelEQ} then yields the corresponding covariance function. 

We first define the Cauchy mixture density as
\begin{equation}
\label{eq:dstCauchy}
\phi_{c}(\bm{s}) = \sum_{q=1}^{Q_c} w_q^c\mathcal{C}(\bm{s};\bm{\eta}_{q},\bm{\gamma}_q), \qquad S_c(\bm{s}) = \frac{\phi_c(\bm{s}) + \phi_c(-\bm{s})}{2},
\end{equation}
where $S_c(\bm{s})$ is symmetric by construction and therefore defines a valid stationary kernel through Bochner’s theorem. 

Substituting $S_c(\bm{s})$ into the inverse Fourier representation, we obtain
\begin{align}
k_{c}(\bm{\tau})
&= \int_{\mathbb{R}^d} S_c(\bm{s})\, e^{2\pi i \bm{s}^{\top}\bm{\tau}}\, d\bm{s} \nonumber\\
&= \frac{1}{2}\int_{\mathbb{R}^d}\!\big(\phi_c(\bm{s})+\phi_c(-\bm{s})\big)\, e^{2\pi i \bm{s}^{\top}\bm{\tau}}\, d\bm{s} \nonumber\\
&=
\frac{1}{2}
\sum_{q=1}^{Q_c} w_q^c
\int_{\mathbb{R}^d}
\mathcal{C}(\bm{s};\bm{\eta}_q,\bm{\gamma}_q)
e^{2\pi i \bm{s}^\top\bm{\tau}}\,d\bm{s}
+
\frac{1}{2}
\sum_{q=1}^{Q_c} w_q^c
\int_{\mathbb{R}^d}
\mathcal{C}(\bm{s};-\bm{\eta}_q,\bm{\gamma}_q)
e^{2\pi i \bm{s}^\top\bm{\tau}}\,d\bm{s}
\nonumber\\
&=
\sum_{q=1}^{Q_c}
w_q^c
e^{-2\pi\,\bm{\gamma}_q^\top |\bm{\tau}|}
\cos\!\left(2\pi\,\bm{\eta}_q^\top \bm{\tau}\right),
\end{align}
where $|\bm{\tau}| := (|\tau_1|,\dots,|\tau_d|)$, $Q_c$ is the number of Cauchy components, $w_q^{c}$ is the corresponding weight.
\subsubsection{Gaussian and Cauchy Spectral Mixture Kernel}
To leverage the complementary properties of both distributions, we define a spectral density $S_{gc}(s)$ as a mixture of Gaussian and Cauchy components:
\begin{equation}
\label{eq:hybrid_density}
S_{gc}(\bm{s})
=
S_g(\bm{s}) + S_c(\bm{s})
=
\frac{1}{2}\bigl(\phi_g(\bm{s})+\phi_g(-\bm{s})\bigr)
+
\frac{1}{2}\bigl(\phi_c(\bm{s})+\phi_c(-\bm{s})\bigr),
\end{equation}
The resulting Gaussian-Cauchy Spectral Mixture (GSM+CSM) then follows: 
\begin{equation}
\label{eq:cont-kernel}
k_{gc}(\bm{\tau}) = \sum_{q=1}^{Q_g} w_q^g e^{-2\pi^2 \bm{\tau}^\top \bm{\Sigma}_q \bm{\tau}} \cos\left(2\pi \bm{\tau}^\top \bm{\mu}_q\right) + \sum_{q=1}^{Q_c}
w_q^c
e^{-2\pi\,\bm{\gamma}_q^\top |\bm{\tau}|}
\cos\!\left(2\pi\,\bm{\eta}_q^\top \bm{\tau}\right).
\end{equation}
First, it maintains spectral interpretability, where each component's location parameters ($\bm{\mu}_q$ and $\bm{\eta}_{q}$) correspond to distinct frequency modes and their weights represent relative energy contributions. Second, it achieves adaptive smoothness through multi-scale modeling capability.

The Gaussian components capture smooth global trends via their bandwidth parameters $\bf{\Sigma}_q$ while the Cauchy components model local variations through their heavy-tailed distributions controlled by $\bm{\gamma}_q$. In other words, the Gaussian and Cauchy mixtures allow the kernel to simultaneously model locally smooth and globally correlated structures in the data. A detailed theoretical derivation and analysis of the proposed method is provided in Supplementary Note 3. In particular, when restricted to the $d$-dimensional continuous input space, we derive upper bounds on the maximum information gain for the GSM and CSM kernels, given by
\begin{equation*}
 \Gamma_{g}(T) = \mathcal{O}\!\left((\log T)^{d+1}\right),\qquad \Gamma_c(T)
=\mathcal{O}\Big(T^{\frac{d+1}{d+2}}(\log T)^{2d-1}\Big) ,\quad
\end{equation*}
respectively, and further establish the corresponding cumulative regret bounds,
\begin{equation*}
R_T^g = \mathcal{O} \left(  (\log T)^{\frac{d+1}{2}} \cdot \sqrt{dT} \right),\qquad R_T^c = \mathcal{O} \left( T^{\frac{2d+3}{2(d+2)}} \sqrt{(\log T)^{2d+1} \cdot d} \right),
\end{equation*}
where $T$ is the iteration number and $d$ is dimension. These results provide formal guarantees on the learning efficiency of the proposed kernels. In addition, we discuss the expressive power of the spectral mixture formulation and show that it can approximate a broad class of kernels (Supplementary Note 3.4), which is corroborated by its superior empirical performance in continuous search spaces compared to conventional kernel choices.

\subsection{Categorical Search Space}
\label{ssex:mixed_kernel}
In addition to purely continuous problems, our spectral mixture kernel also generalizes to mixed categorical-continuous spaces, a setting frequently encountered in chemical experimentation but hitherto underexplored in the literature. 
 
For categorical inputs, we adopt weighted Hamming similarity and perform acquisition optimization using the trust-region alternating algorithm~\cite{ru2020bayesian,wan2021think,eriksson2019scalable}. 
Let $\bm{x}_u, \bm{x}_u' \in \mathcal{X}_u $
denote two $U$-dimensional categorical inputs. The resulting kernel is defined as
\begin{equation}
\label{eq:cate-kernel}
k_u(\bm{x}_u, \bm{x}_u') =
\exp\!\left(
    \frac{1}{U}
    \sum_{i=1}^{U}
    \ell_i \, \delta\!\left(x_u^{(i)}, x_u'^{(i)}\right)
\right),
\end{equation}
where $\delta(\cdot,\cdot)$ denotes the Kronecker delta function,  $x_u^{(i)}$ denotes the $i$-th categorical component, and
$\{\ell_i\}_{i=1}^{U}$ are lengthscale parameters, which are learned from data by maximizing the MLL.

Following~\cite{ru2020bayesian}, we construct a composite kernel by combining the spectral mixture kernel for continuous variables and the Hamming kernel for categorical variables: 
\begin{equation}
\label{eq:comp-kernel}
k(\bm{x}, \bm{x}') =
\lambda \Big(k_{gc}(\bm{x}_c, \bm{x}_c')\, k_u(\bm{x}_u, \bm{x}_u')\Big)+ (1 - \lambda) \Big( k_{gc}(\bm{x}_c, \bm{x}_c') + k_u(\bm{x}_u, \bm{x}_u')\Big),
\end{equation}
where $\lambda \in [0, 1]$ is a trade-off parameter. This kernel is a convex combination of a multiplicative component and an additive component. The product term couples continuous and categorical similarities, allowing the response surface over continuous variables to vary across different categorical configurations and thereby capturing cross-variable interactions. The additive term models independent contributions from continuous and categorical inputs, enabling information sharing even when one component differs. The parameter $\lambda$ balances these two effects, controlling the degree of interaction between continuous and categorical dimensions.

\section{Code availability}
Code and an interactive website for this work are publicly available at GitHub:  \href{https://github.com/Lab-of-Catalysis-AI/CatBOX}{https://github.com/Lab-of-Catalysis-AI/CatBOX} and at \href{https://catbox.top}{https://catbox.top}.

\clearpage

\bibliography{main}

\section{Acknowledgements}
This work is supported by National Key R\&D Program of China (2024YFA1509901),  National Natural Science Foundation of China (72301172, 72394370:72394375, 72495130:72495132), Shanghai Education Commission Chenguang Program (22CGA12), and Shanghai Jiao Tong University Office of Liberal Arts (ZHWK2502).
\section{Competing interests}
The authors declare a pending patent related to this work.
\section{Author contributions}
C.Z.: Methodology, Website, Validation, Investigation, Data curation, Writing of original draft, Review and editing, Visualization. Y.Z.: Conceptualization, Methodology, Validation, Investigation, Data curation, Writing of original draft, Review and editing. Z.L. and L.J.: Review and editing. C.H. and Y.H.: Conceptualization, Methodology, Supervision, Funding Acquisition, Writing of original draft, Review and editing.

\end{document}